\documentclass{fairmeta}
\microtypesetup{expansion=false}

\usepackage{amsmath}
\usepackage{amssymb}
\usepackage{amsfonts}
\usepackage{enumitem}

\newcommand{\method}{LOCI}
\newcommand{\flux}{\textsc{Flux.1-Kontext}}
\newcommand{\apita}{IDAttn}

\newcommand{\real}{\mathbb{R}}

\usepackage{tikz}

\definecolor{sub1}{RGB}{214,39,40}
\definecolor{sub2}{RGB}{31,119,180}
\definecolor{sub3}{RGB}{44,160,44}
\definecolor{sub4}{RGB}{255,127,14}
\tikzset{boxA/.style={draw=sub1,line width=0.6pt},
         boxB/.style={draw=sub2,line width=0.6pt},
         boxC/.style={draw=sub3,line width=0.6pt},
         boxD/.style={draw=sub4,line width=0.6pt}}
\newcommand{\flame}{\raisebox{-0.15ex}{\includegraphics[height=0.85em]{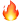}}}

\newcommand{\qsrc}[2]{%
  \begin{tikzpicture}
    \node[anchor=south west,inner sep=0] (im) at (0,0)
      {\includegraphics[width=#2]{figures/qual/#1/src.jpg}};
    \begin{scope}[x={(im.south east)},y={(im.north west)}]
      \csname qbb#1\endcsname
    \end{scope}
  \end{tikzpicture}}

\newlength{\qw}
\newlength{\qin}
\newcommand{\qim}[2]{\includegraphics[width=\qw]{figures/qual/#1/#2.jpg}}
\newcommand{\qlab}[2]{\parbox[b][\qw][c]{1.8ex}{\centering%
  \rotatebox{90}{\tiny\color{#1}#2}}}
\newcommand{\qualblock}[2]{%
  \begin{minipage}[c]{\qin}%
    \qsrc{#1}{\qin}\\[1pt]%
    {\centering\tiny Input\par}\vspace{2.5pt}%
    {\raggedright\tiny #2\par}%
  \end{minipage}\hspace{2pt}%
  \begin{minipage}[c]{4.81cm}%
    \setlength{\tabcolsep}{0.8pt}\renewcommand{\arraystretch}{0.3}%
    \begin{tabular}{@{}c@{\hspace{1.5pt}}ccc@{}}
      & \multicolumn{3}{c}{\tiny stronger edit $\longrightarrow$}\\
      \qlab{sub1}{Subject 1} & \qim{#1}{A2} & \qim{#1}{A3} & \qim{#1}{A4}\\
      \qlab{sub2}{Subject 2} & \qim{#1}{B2} & \qim{#1}{B3} & \qim{#1}{B4}\\
      \qlab{black}{Both}     & \qim{#1}{C2} & \qim{#1}{C3} & \qim{#1}{C4}\\
    \end{tabular}%
  \end{minipage}}
\newcommand{\qualpair}[4]{\qualblock{#1}{#2}\hfill\qualblock{#3}{#4}}

\newlength{\gw}
\newcommand{\gim}[2]{\includegraphics[width=\gw]{figures/qual/#1/#2.jpg}}
\newcommand{\glab}[2]{\parbox[b][\gw][c]{2ex}{\centering%
  \rotatebox{90}{\scriptsize\color{#1}#2}}}
\newcommand{\qualquad}[3][Subject]{%
  \begin{minipage}[c]{4.3cm}%
    \qsrc{#2}{4.3cm}\\[1pt]%
    {\centering\scriptsize Input\par}\vspace{3pt}%
    {\raggedright\tiny #3\par}%
  \end{minipage}\hspace{4pt}%
  \begin{minipage}[c]{11.4cm}%
    \setlength{\tabcolsep}{1pt}\renewcommand{\arraystretch}{0.3}%
    \begin{tabular}{@{}c@{\hspace{2pt}}cccc@{}}
      & \multicolumn{4}{c}{\scriptsize edit strength increases
        ($\alpha$ decreasing) $\longrightarrow$}\\
      \glab{sub1}{#1 1} & \gim{#2}{A1} & \gim{#2}{A2} & \gim{#2}{A3} & \gim{#2}{A4}\\
      \glab{sub2}{#1 2} & \gim{#2}{B1} & \gim{#2}{B2} & \gim{#2}{B3} & \gim{#2}{B4}\\
      \glab{sub3}{#1 3} & \gim{#2}{C1} & \gim{#2}{C2} & \gim{#2}{C3} & \gim{#2}{C4}\\
      \glab{sub4}{#1 4} & \gim{#2}{D1} & \gim{#2}{D2} & \gim{#2}{D3} & \gim{#2}{D4}\\
      \glab{black}{All} & \gim{#2}{E1} & \gim{#2}{E2} & \gim{#2}{E3} & \gim{#2}{E4}\\
    \end{tabular}%
  \end{minipage}}

\newlength{\tw}
\newcommand{\tim}[2]{\includegraphics[width=\tw]{figures/qual/#1/#2.jpg}}
\newcommand{\tlab}[2]{\parbox[b][\tw][c]{2ex}{\centering%
  \rotatebox{90}{\scriptsize\color{#1}#2}}}
\newcommand{\qualteaser}[2]{%
  \begin{minipage}[c]{2.9cm}%
    \qsrc{#1}{2.9cm}\\[1pt]%
    {\centering\scriptsize Input\par}\vspace{3pt}%
    {\raggedright\tiny #2\par}%
  \end{minipage}\hspace{3pt}%
  \begin{minipage}[c]{13.2cm}%
    \setlength{\tabcolsep}{1pt}\renewcommand{\arraystretch}{0.3}%
    \begin{tabular}{@{}c@{\hspace{2pt}}cccc@{}}
      & \multicolumn{4}{c}{\scriptsize edit strength increases
        ($\alpha$ decreasing) $\longrightarrow$}\\
      \tlab{sub1}{Subject 1} & \tim{#1}{A2} & \tim{#1}{A3} & \tim{#1}{A4} & \tim{#1}{A5}\\
      \tlab{sub2}{Subject 2} & \tim{#1}{B2} & \tim{#1}{B3} & \tim{#1}{B4} & \tim{#1}{B5}\\
      \tlab{black}{Both}     & \tim{#1}{C2} & \tim{#1}{C3} & \tim{#1}{C4} & \tim{#1}{C5}\\
    \end{tabular}%
  \end{minipage}}

\title{LOCI: Localized Continuous Control for Multi-Subject Image Editing}

\author[1]{Sai Vidyaranya Nuthalapati}
\author[1]{Shlok Kumar Mishra}

\affiliation[1]{Meta AI}

\abstract{%
Recent instruction-following image editors such as \flux{} enable text-driven
editing but lack fine-grained control over \emph{where} and \emph{how much} an
edit is applied.  Existing approaches provide either continuous but global
control or spatially localized editing at a fixed strength.  To our knowledge,
none jointly supports explicit spatial localization, continuous per-region
strength, and independent simultaneous control of multiple subjects.  We
present \method{} (\textbf{LO}calized \textbf{C}ontinuous control for
multi-subject \textbf{I}mage editing).  Given a multi-subject image with
per-subject regions and instructions, \method{} edits each subject at an
independently chosen, continuously adjustable strength, for instance gradually
aging one person while adding a beard to another with minimal cross-region
leakage.  \method{} pairs subject-level attention binding for spatial isolation
with a \emph{Regional LoRA} layer that replaces the LoRA's scalar coefficient
with a per-token scaling vector aligned to the spatial layout.  A single
adapter is trained at one suppression point, and continuous edit strengths
follow by rescaling its regional residual at inference.  Experiments on
multi-subject scenes demonstrate that \method{} achieves strong edit fidelity
with significantly reduced cross-region interference compared to baselines,
while maintaining continuous control over edit strength.}

\date{\today}
\correspondence{ \email{saividyaranya@meta.com}}

\begin{document}

\maketitle

\begin{figure}[h!]
  \centering
  \qualteaser{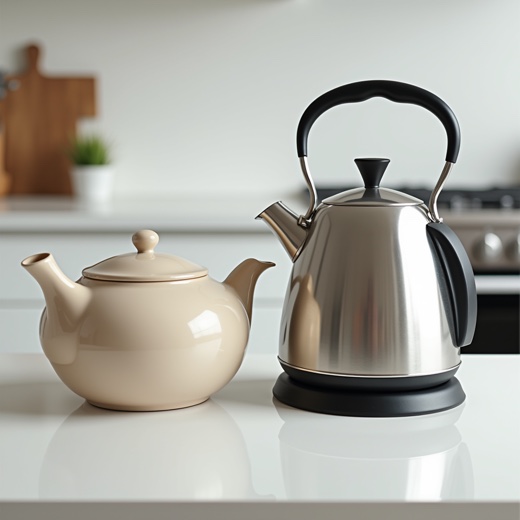}{%
    {\color{sub1}\emph{transform the teapot into an ornate steampunk
    contraption with brass pipes, pressure gauges and copper rivets}}\par
    \vspace{2pt}%
    {\color{sub2}\emph{change the teapot into a vessel carved from luminous
    jade, with dragon motifs and a serpent handle}}}
  \caption{\textbf{\method{} gives every subject its own continuous edit
    control.}  From one input, the top row strengthens the edit on the first
    teapot while the second stays pinned at the neutral anchor, the middle row
    does the converse, and the bottom row moves both controls together.  The
    boxes on the input are the spatial annotations supplied to the model:
    {\color{sub1}red} for subject~1 and {\color{sub2}blue} for subject~2,
    matching the row labels.}
  \label{fig:teaser}
\end{figure}

% !TeX root = ../main.tex
\section{Introduction}
\label{sec:intro}

% \begin{figure*}[t]
%   \centering
%   \includegraphics[width=\textwidth]{figures/teaser.pdf}
%   \caption{\textbf{\method{}.} Given a multi-subject image, our method enables independent, continuously adjustable edits per subject. Here, one person is gradually aged ($\alpha{=}{-}1$) while the other receives a beard ($\alpha{=}{-}0.5$), each controlled by an independent slider with smooth interpolation. Recall that $\alpha{=}1$ leaves a subject unedited and edit strength increases as $\alpha$ decreases. No existing method provides this combination of spatial control, continuous strength, and per-region independence.}
%   \label{fig:teaser}
% \end{figure*}

The rise of text-to-image (T2I) generative models~\citep{ramesh2022dalle2, saharia2022imagen, podell2023sdxl, rombach2022high} has revolutionized image editing,
enabling creative expression at an unprecedented scale through simple textual prompts.
Ranging from Generative Adversarial Networks ~\citep{goodfellow2014generative} to diffusion models~\citep{sohl2015deep, song2020score, dhariwal2021diffusion},
to flow-based models~\citep{lipman2022flow, liu2022rectified, esser2024sd3, flux}, these models have demonstrated the ability to generate high-quality images which adhere to the text prompts.
Recently diffusion and flow matching models have excelled at various tasks such as inpainting~\citep{lugmayr2022repaint},
outpainting~\citep{saharia2022palette}, style transfer~\citep{zhang2023inst}, and instruction-based editing~\citep{brooks2023instructpix2pix, zhang2024magicbrush, sheynin2024emu, flux_kontext} to name a few.

Existing editors fall into two families. Firstly, continuous editors expose an edit-strength control but reach a subject only through the prompt~\citep{slideredit, kontinuouskontext}.
Language grounding is coarse: it can express ``replace the dog with a cat,'' but not ``replace the second dog from the left'' when several same-category instances co-occur.
Because the control is never bound to an image region, it cannot designate \emph{which} subject to edit,
cannot assign independently chosen intensities to several subjects in one pass~\citep{gandikota2023concept, ye2025allinone, endo2026flowslider},
and cannot confine the edit once it is applied without spilling onto its neighbours where an edit intended for one subject spills onto its neighbours ~\citep{dahary2024bounded, mun2025leakage}.

The second family binds edits to regions, through masks or attention, and so resolves both designation and containment.
These methods, however, apply the edit at a single fixed strength,
offering no interpolation between the source image and the fully applied edit~\citep{lomoe, idattn, cai2026editwhereyoumean, xiong2026bridge}.

The two capabilities therefore live in separate families.
To our knowledge, no existing approach combines explicit regional localization,
continuous per-region edit strength, and independent simultaneous control of multiple subjects in a single editing process.

We present \method{}, a single framework that resolves these limitations.
Given a multi-subject image annotated with per-subject regions and edit instructions, \method{} edits each designated subject at an independently chosen,
continuously adjustable strength while confining every edit to its own region.
Each region is given as either a bounding box or a segmentation mask, and it supplies the explicit spatial addressing that language grounding lacks, so the user can specify exactly which subject to modify.
Scaling each region's contribution independently gives every subject its own continuous strength within a single forward pass,
and attention masking isolates the regions to suppress cross-subject leakage.

Realizing per-region strength control is nontrivial.
The conditioning paths in MMDiT backbones~\citep{esser2024sd3, flux} (AdaLN modulation and attention) govern \emph{what} features are computed,
not \emph{how intensely} an edit is applied; modulating their parameters does not yield a linear interpolation between unedited and edited outputs.
Attention masking can partition which edit reaches which region~\citep{idattn}, but this isolation is binary and carries no notion of graded strength.
On the other hand, the low-rank update of LoRA~\citep{hu2022lora}, enters as an additive residual whose magnitude directly governs edit intensity.
Scaling it by zero recovers the base output, scaling by one applies the learned update in full, and intermediate values interpolate continuously by construction~\citep{slideredit}.
However, scaling the residual per region is not sufficient on its own because the learned edit direction is spatially entangled:
the residual is written only inside a region, but attention carries its effect outward, so assigning different scales to
different regions does not produce independent edits.
Our key insight is that training the LoRA under instance-aware attention masking yields region-local edit directions.
Once each direction is local, per-region scaling of the residual becomes meaningful, and each subject can be edited at an independently chosen strength within a single forward pass,
without modifying the base architecture.
We make the following contributions.

\begin{enumerate}[nosep,leftmargin=*]
\item We introduce \textbf{\method{}}, a framework for independently controllable multi-subject editing
  that represents an edit as an instruction, a spatial support, and a continuous control coefficient.
  At its core is \textbf{Regional LoRA}, a layer that scales the residual separately in each region instead of applying
  one strength to the whole image, so every subject carries its own control and the background carries none.
  Multiple subject-specific edits are then composed within one model while their strength and spatial influence stay independently controlled.

\item We curate a \textbf{multi-subject evaluation set} comprising 128 scenes and 427 source--region--edit tuples spanning people,
  animals, vehicles, furniture, and mixed compositions with 2--7 subjects per scene, and define four evaluation protocols (single-target editing,
  independent multi-control sweeps, joint editing, and two-dimensional Cartesian grid control) that together probe target fidelity,
  cross-subject leakage, continuity, and slider independence.
\end{enumerate}

% !TeX root = ../main.tex
\section{Related Work}
\label{sec:related}

\paragraph{Instruction-following image editing.}
Instruction-following editing has advanced rapidly through better training
data and multi-task objectives~\citep{brooks2023instructpix2pix,
zhang2024magicbrush, sheynin2024emu}.
We build on \flux{}~\citep{flux_kontext}, a prominent recent architecture that
represents the source image as latent tokens concatenated into the
transformer's sequence, a property our construction relies on
(\cref{sec:method_prelim}).

\paragraph{Spatial control.}
Attention manipulation localizes edits without training~\citep{hertz2022prompt, cao2023masactrl},
and layout-guided generation injects boxes or keypoints as conditions~\citep{li2023gligen}.
For multi-instance editing, Zaccagnino et al.~\citep{idattn} build disentanglement and harmonization masks
from box layouts to block cross-instance leakage, which we adopt as our spatial backbone.
Masked inpainting~\citep{rombach2022high} localizes an edit but needs a binary region and offers no graded strength.
None of these vary edit intensity smoothly.

\paragraph{Continuous edit-strength control.}
Concept Sliders~\citep{gandikota2023concept} learn attribute directions in LoRA weight space.
SliderEdit~\citep{slideredit} scales instruction-indexed token-selective adapters,
and Kontinuous Kontext~\citep{kontinuouskontext} maps a scalar strength to instruction-conditioned global modulation.
Both reach a subject only through language,
so neither binds a strength control to a user-specified image region.

\paragraph{Multi-subject and localized editing.}
Mask-guided multi-object editing~\citep{lomoe} and regional LoRA composition~\citep{dalva2025lorashop}
edit several subjects at once but at a fixed intensity,
as do patch-weighted concept merging~\citep{foteinopoulou2026cds}
and region-aware adapters~\citep{cai2026editwhereyoumean, xiong2026bridge}.
Layered Diffusion Brushes~\citep{layeredbrushes} exposes a regeneration amount tied to the noise injection timestep.
Neither applies distinct per-subject edits, and neither isolates regions at the attention level.
\Cref{sec:supp_related} expands this comparison.

\paragraph{Positioning.}
Existing work addresses continuous strength, spatial control, or multi-subject isolation separately.
\method{} unifies region-based addressing, per-token LoRA scaling, and instance-disentangled attention,
so several subjects can be edited at independently chosen strengths in a single forward pass.

% !TeX root = ../main.tex
\section{Method}
\label{sec:method}

\begin{figure*}[t]
  \centering
  \includegraphics[width=\textwidth]{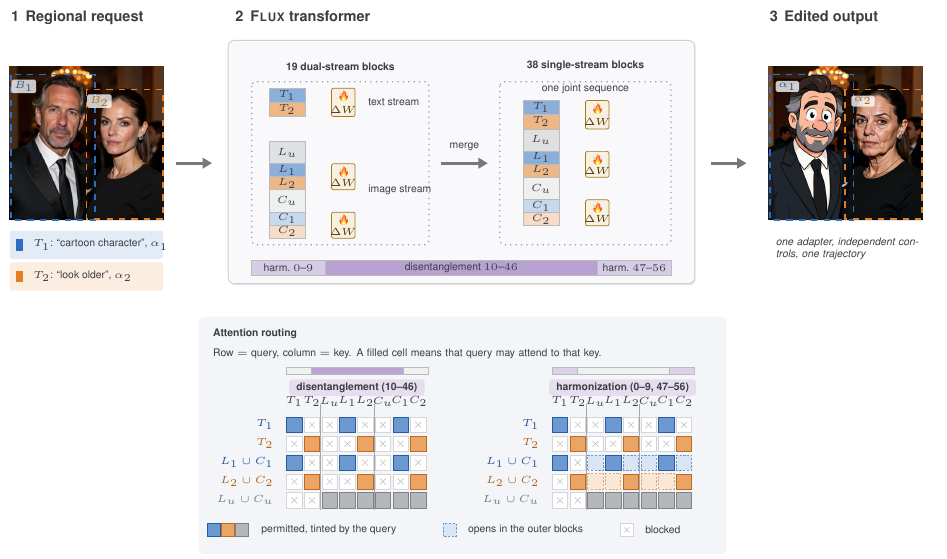}
  \caption{\textbf{Architecture of \method{}.}
    \emph{Stage 1}: the user supplies a source image, per-region bounding
    boxes $B_n$, edit instructions, and independent scalar controls
    $\alpha_n$.  \emph{Stage 2}: text and image tokens are organized into the
    partition $\{T_n, L_n, C_n\}$ indexed by subject, with trainable low-rank
    adapters ($\Delta W$, marked with \flame) whose residuals are gated per
    token by $\alpha_n$; background tokens $L_u, C_u$ receive zero adapter
    contribution.  \emph{Stage 3}: the frozen \flux{} transformer processes the
    sequence through 19 dual-stream and 38 single-stream blocks under a
    three-zone attention schedule, with harmonization in the outer blocks
    (0--9, 47--56) and disentanglement in the middle blocks (10--46).
    \emph{Inset}: the query--key attention matrices for both regimes.  Under
    disentanglement, each subject's tokens attend only
    within their own partition; under harmonization,
    subject image tokens additionally read all visual tokens (dashed cells),
    restoring the cross-region evidence needed for background
    coherence.}
  \label{fig:architecture}
\end{figure*}

A multi-subject editing request consists of a source image $I \in \real^{H \times W \times 3}$
and $N$ region-specific instructions $\{(e_n, M_n)\}_{n=1}^{N}$,
where $e_n$ is the instruction for subject $n$ and $M_n \in \{0,1\}^{H \times W}$ is its binary region mask.
The user supplies each region as either a bounding box or a segmentation mask.
A segmentation mask is used directly, and a bounding box is converted by marking every pixel inside it as foreground,
so the method takes a single region representation $M_n$ in both cases.
For attention routing, we additionally associate each region with a bounding box. This can be the supplied box itself or the tight box enclosing a supplied segmentation mask.
Each subject also has a control value $\alpha_n \in \real$, the slider exposed to the user.
Our goal is to produce an edited image in which every subject realizes its own instruction
at the strength set by its own control, while the background is preserved.
We use a suppression parameterization, in which $\alpha_n = 1$ leaves subject $n$ unedited,
$\alpha_n = 0$ applies the nominal edit, and $\alpha_n < 0$ extrapolates beyond it,
where the \emph{nominal edit} is the output that the frozen \flux{} editor produces for $e_n$ on its own.

\Cref{sec:method_prelim} fixes the backbone and the token partition our construction is built on.
\Cref{sec:method_lora} introduces Regional LoRA, which turns the scalar adapter scale into a per-token vector aligned with that partition.
\Cref{sec:method_apita} describes the attention routing, which binds each instruction to its own subject,
and \cref{sec:method_training} the objective that trains one shared adapter on single-region examples only, under exactly the routing used at test time.
\Cref{sec:method_inference} composes an arbitrary number of controls in a single pass.

\subsection{Preliminaries}
\label{sec:method_prelim}

We first fix the backbone, the token partition that our construction indexes,
and the low-rank adapter that it modifies.

\paragraph{Backbone and token partition.}
We build on the frozen \flux{}~\citep{flux_kontext} editor, an MMDiT~\citep{esser2024sd3} with $19$ dual-stream and $38$ single-stream blocks.
The model is trained under a rectified-flow objective~\citep{lipman2022flow, liu2022rectified, esser2024sd3} and predicts a velocity.
We denote its input token sequence by $Z = (Z_1,\dots,Z_{|Z|})$,
\begin{equation}
  Z = Z^{\text{text}} \,\|\, Z^{\text{target}} \,\|\, Z^{\text{context}} ,
  \label{eq:sequence}
\end{equation}
where $\|$ denotes concatenation along the sequence axis.
The stream $Z^{\text{text}}$ contains the instruction tokens produced by T5~\citep{raffel2020t5}.
The other two streams are latent tokens from the frozen \flux{} autoencoder~\citep{flux_kontext}:
$Z^{\text{target}}$ is the evolving latent that the model denoises into the edited image,
and $Z^{\text{context}}$ is a fixed encoding of the source image, which conditions the edit but is not itself denoised.

For subject-indexed control, we organize the three streams by subject association.
The text stream physically concatenates independently encoded blocks,
$Z^{\text{text}} =  T_1 \| \cdots \| T_N$,
where $T_n$ holds instruction $e_n$.
The image tokens remain in their original sequence order.
For attention routing, we write $L_n \subseteq Z^{\text{target}}$ and $C_n \subseteq Z^{\text{context}}$ for the target and context tokens whose spatial cells lie inside the region mask associated with subject $n$.
The tokens outside all routing boxes form $L_u$ and $C_u$.
We can represent them as
\begin{equation}
  Z^{\text{target}} = L_u \cup \bigcup_{n=1}^{N} L_n,
  \qquad
  Z^{\text{context}} = C_u \cup \bigcup_{n=1}^{N} C_n.
  \label{eq:image_partitions}
\end{equation}
We call this labelling of every token by stream and subject the \emph{token partition}.

\paragraph{Low-rank update:}
For a layer mapping input $x \in \real^{d_{\text{in}}}$ to output $y \in \real^{d_{\text{out}}}$,
LoRA~\citep{hu2022lora} augments the frozen projection $W \in \real^{d_{\text{out}} \times d_{\text{in}}}$
with an update $\Delta W = BA$ of rank at most $r$,
where $A \in \real^{r \times d_{\text{in}}}$ and $B \in \real^{d_{\text{out}} \times r}$ are trainable.
A scalar coefficient $\beta \in \real$ controls the update,
\begin{equation}
  y = Wx + \beta\,\Delta W x .
  \label{eq:lora}
\end{equation}
The update is linear in $\beta$, so $\beta=0$ recovers the frozen projection and $\beta=1$ applies the learned update in full.

\subsection{Regional LoRA}
\label{sec:method_lora}

The coefficient $\beta$ in \cref{eq:lora} applies to all tokens equally.
To give each region its own edit strength,
we replace it with a vector $\boldsymbol{\lambda} \in \real^{|Z|}$ that holds one coefficient $\lambda_j$ per token,
defined over the token partition of \cref{sec:method_prelim}.
To project $M_n$ from pixel space into the token grid, we define
\begin{equation}
  G_n(j) = \max_{p \in c(j)} M_n(p) ,
  \label{eq:mask_projection}
\end{equation}
where $c(j)$ is the block of source pixels that token $j$ covers.
The gate is therefore one whenever any pixel of subject $n$ falls in that block,
and both image streams share the same token grid, so token $j$ covers the same block in $Z^{\text{target}}$ and $Z^{\text{context}}$.
The coefficient at each token is then
\begin{equation}
  \lambda_j =
  \begin{cases}
    \alpha_n                 & Z_j \in T_n,\ n \in \{1,\ldots,N\}, \\
    \alpha_{n^\star(j)}      & G_n(j) = 1 \text{ for some } n, \\
    0                        & \text{otherwise,}
  \end{cases}
  \label{eq:regional_scaling}
\end{equation}
where
$n^\star(j)=\arg\max_{n:\,G_n(j)=1}\alpha_n$.
Thus, when projected gates overlap, an image token receives the largest coefficient among the active controls, that is, the most suppressive and hence least edited of them.
The background tokens $L_u, C_u$ carry $\lambda_j=0$ and receive no adapter residual.
Substituting $\lambda_j$ for the scalar $\beta$ of \cref{eq:lora}, each LoRA layer computes
\begin{equation}
  y_j = W x_j + \lambda_j\,\Delta W x_j .
  \label{eq:regional_lora}
\end{equation}
Thus $\alpha_n$ modulates three groups of low-rank writes, namely its instruction block $T_n$,
its projected region in the target stream, and the corresponding region in the context stream.

\begin{figure*}[!t]
  \centering
  \qualpair{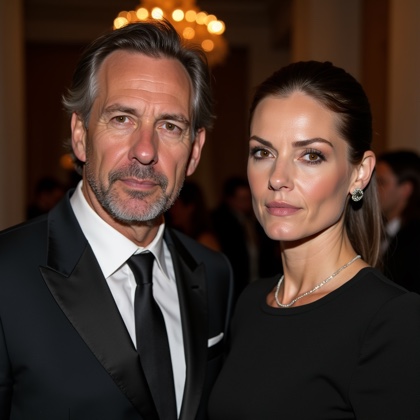}{%
    {\color{sub1}\emph{turn the person into a cartoon character}}\par\vspace{2pt}%
    {\color{sub2}\emph{make the person look older}}}%
  {interview}{%
    {\color{sub1}\emph{make the person laugh}}\par\vspace{2pt}%
    {\color{sub2}\emph{turn the person into a metallic figure}}}\\[7pt]
  \qualpair{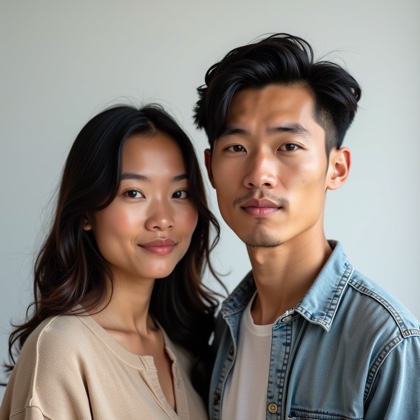}{%
    {\color{sub1}\emph{add makeup to the person}}\par\vspace{2pt}%
    {\color{sub2}\emph{make the person laugh}}}%
  {wall}{\emph{one instruction, both regions: change the colour of the
    climbing wall from gray to a tan shade}}
  \caption{\textbf{Independent and joint control across four scenes.}
    Each block shows its input and instructions on the left, with
    {\color{sub1}red} and {\color{sub2}blue} boxes marking the spatial
    annotations given to the model.  The three rows sweep subject~1,
    subject~2, and both, and within a row the edit strengthens from left to
    right.
    see \cref{sec:qual_results}.}
  \label{fig:qual}
\end{figure*}

\subsection{Instance-Disentangled Attention}
\label{sec:method_apita}

The write gate of \cref{sec:method_lora} localizes where the residual is added, but not where its effect lands: attention lets a change at one token reach every other token in the following blocks.
We therefore route attention with Instance-Disentangled Attention (\apita{})~\citep{idattn}, which adds a bias $\Omega \in \{0,-\infty\}^{|Z| \times |Z|}$ inside the softmax,
\begin{equation}
  \operatorname{\apita{}}(Q, K, V, \Omega) = \mathrm{softmax}\!\left(\tfrac{QK^\top}{\sqrt{d}} + \Omega\right) V ,
  \label{eq:apita}
\end{equation}
where $Q$, $K$ and $V$ are the query, key and value projections of $Z$ and $d$ is the per-head dimension.

Two regimes are defined on the partition.
Under the \emph{disentanglement} regime $\Omega^{\text{dis}}$, the three blocks of subject $n$, namely $T_n$, $L_n$ and $C_n$, attend only among themselves,
while background queries in $L_u$ and $C_u$ may read every target and context token.
The relation is therefore asymmetric.
Further, no query outside subject $n$ may read its instruction block $T_n$.
The \emph{harmonization} regime $\Omega^{\text{har}}$ modifies only the subject rows,
letting $L_n$ and $C_n$ read all target and context tokens,
which restores the cross-region evidence that lighting, shadow and background continuity depend on
while leaving instruction binding unchanged ~\citep{idattn}.
We apply $\Omega^{\text{har}}$ in the outer blocks $0$ to $9$ and $47$ to $56$,
and $\Omega^{\text{dis}}$ in the middle blocks $10$ to $46$,
using the same schedule at training and inference.

\begin{table*}[!t]
  \centering
  \scriptsize
  \setlength{\tabcolsep}{3.5pt}
  \begin{tabular}{@{}clcccccccccc@{}}
    \toprule
    & & \multicolumn{2}{c}{Extrapolation $\uparrow$}
    & \multicolumn{2}{c}{Direction corr.\ $\downarrow$}
    & \multicolumn{2}{c}{Mean range $\uparrow$}
    & \multicolumn{4}{c}{Disentanglement $\downarrow$} \\
    \cmidrule(lr){3-4}\cmidrule(lr){5-6}\cmidrule(lr){7-8}\cmidrule(l){9-12}
    Protocol & Method & CLIP & SigLIP & CLIP & SigLIP & CLIP & SigLIP
      & LPIPS-Alex & LPIPS-VGG & DINOv2 & Identity \\
    \midrule
    A & (1) SliderEdit
      & 0.2431 & 0.1114 & $-0.4907$ & $-0.4666$ & 0.0575 & 0.0470
      & 0.3315 & 0.3826 & 0.1612 & 0.3640 \\
    A & (2) SliderEdit $+$ \apita{}
      & \underline{0.2500} & \underline{0.1187} & \underline{$-0.6326$} & \underline{$-0.6192$} & 0.0726 & 0.0587
      & \underline{0.1508} & \underline{0.2010} & \underline{0.0609} & \underline{0.2598} \\
    A & (4) Regional LoRA without \apita{}
      & 0.2445 & 0.1133 & $-0.5829$ & $-0.5689$ & \underline{0.0761} & \underline{0.0625}
      & 0.3085 & 0.3581 & 0.1700 & 0.3406 \\
    A & (5) \method{} (ours)
      & \textbf{0.2527} & \textbf{0.1207} & $\mathbf{-0.7081}$ & $\mathbf{-0.6953}$ & \textbf{0.0894} & \textbf{0.0721}
      & \textbf{0.1467} & \textbf{0.1979} & \textbf{0.0550} & \textbf{0.2548} \\
    \addlinespace
    B & (1) SliderEdit
      & 0.2426 & 0.1121 & $-0.5503$ & $-0.5106$ & 0.0589 & 0.0488
      & 0.3346 & 0.3875 & 0.1471 & 0.3502 \\
    B & (2) SliderEdit $+$ \apita{}
      & \underline{0.2479} & \underline{0.1181} & $-0.6199$ & $-0.6284$ & 0.0708 & 0.0578
      & \underline{0.1128} & \underline{0.1500} & \underline{0.0804} & \underline{0.2397} \\
    B & (4) Regional LoRA without \apita{}
      & 0.2439 & 0.1131 & \underline{$-0.6636$} & \underline{$-0.6438$} & \underline{0.0807} & \underline{0.0649}
      & 0.2331 & 0.2835 & 0.1176 & 0.3088 \\
    B & (5) \method{} (ours)
      & \textbf{0.2531} & \textbf{0.1221} & $\mathbf{-0.7078}$ & $\mathbf{-0.6977}$ & \textbf{0.0890} & \textbf{0.0720}
      & \textbf{0.0748} & \textbf{0.1128} & \textbf{0.0508} & \textbf{0.1988} \\
    \bottomrule
  \end{tabular}
  \caption{\textbf{Protocols~A and B.}  Best per column \textbf{bold},
    second best \underline{underlined}, ranked within each protocol
    block.}
  \label{tab:slideredit_metrics_available}
\end{table*}

\subsection{Single-Region Suppression Training}
\label{sec:method_training}

\Cref{sec:method_lora,sec:method_apita} fix how the adapter is applied and how attention is routed. Here we describe how it is trained.
The frozen editor supplies its own supervision.
Run on a source image with a neutral instruction it reproduces that image, and run with the real instruction it produces the edit.
Following~\citep{slideredit}, we train a single low-rank adapter, one subject at a time, to turn the second prediction into the first,
so that at $\alpha=1$ it removes exactly the change the instruction causes.
No edited image is therefore required: each example is a source image $I$, an instruction $e$, and its region $M$.
Training uses the same attention routing as inference.

\paragraph{Routing-matched suppression.}
For each training tuple, we encode $I$ once.
One copy of its latent serves as the clean context latent $X_{\mathrm{src}}$,
while another is corrupted with noise by the standard rectified-flow interpolation to form the denoising latent $X_t$ at a logit-normally sampled noise level~\citep{liu2022rectified,esser2024sd3}.
We denote by $\mathcal{F}_{\mathcal{R}}$ the target-stream velocity that \flux{} predicts under a routing schedule $\mathcal{R}$,
and by $\mathcal{F}^{\boldsymbol{\lambda}}_{\mathcal{R}}$ the same prediction with the adapter active at scale vector $\boldsymbol{\lambda}$.
The two passes share these image-derived latents, the sampled noise level, token layout, region and routing schedule,
but differ in their instruction conditioning and adapter state:
\begin{equation}
\begin{aligned}
  v_{\mathrm{ref}}   &= \mathcal{F}_{\mathcal{R}}\bigl(X_t,\, X_{\mathrm{src}},\, e_{\mathrm{neu}}\bigr), \\
  v_{\mathrm{adapt}} &= \mathcal{F}^{\boldsymbol{\lambda}}_{\mathcal{R}}\bigl(X_t,\, X_{\mathrm{src}},\, e\bigr),
\end{aligned}
\label{eq:two_passes}
\end{equation}
where $e_{\mathrm{neu}}$ is the neutral instruction \emph{``keep the image the same''},
$\boldsymbol{\lambda}$ follows from $\alpha=1$ and the gate $G_1$ of region $M$ by \cref{eq:regional_scaling},
and $\mathcal{R}$ is the schedule of \cref{sec:method_apita}.
The detached $v_{\mathrm{ref}}$ is the velocity target for optimization, and gradients update the adapter through $v_{\mathrm{adapt}}$.
The instruction block $T_1$ is the only conditioning that differs between the two velocity calculations, and the routing schedule in particular is shared.
If the reference pass instead used relaxed attention, the loss would ask the adapter to do two things at once:
suppress the instruction and compensate for the difference between relaxed attention and \apita{} routing.
At inference only the \apita{} schedule is present, yet the regional control would rescale both learned effects together.
Using the same schedule in both passes therefore isolates edit suppression at a fixed routing operating point.

\paragraph{Masked objective.}
Let $m \in \{0,1\}^{|Z^{\text{target}}|}$ be the binary training mask derived from the token-space gate $G_1$ of \cref{eq:mask_projection}.
We supervise only the corresponding target-stream tokens:
\begin{equation}
  \mathcal{L} =
  \frac{\sum_{j=1}^{|Z^{\text{target}}|} m_j
  \,\bigl\|v_{\mathrm{adapt},j}-v_{\mathrm{ref},j}\bigr\|_2^2}
  {\sum_{j=1}^{|Z^{\text{target}}|}m_j}.
  \label{eq:masked_loss}
\end{equation}

\subsection{Multi-Region Inference}
\label{sec:method_inference}

Although training uses a single subject, inference may instantiate $N>1$ controls on the token partition of \cref{sec:method_prelim}.
We assemble all regional coefficients into a single vector $\boldsymbol{\lambda}$ using \cref{eq:regional_scaling} which modulates the shared adapter at every denoising step.
The controls may have different signs, allowing suppression and extrapolation to be applied to different subjects in the same trajectory.

Inference applies the support pattern used during single-region training to each subject,
so that its instruction block $T_n$ and the image tokens selected by $G_n$ share the coefficient $\alpha_n$.
On subject $n$'s tokens, $\boldsymbol{\lambda}$ is therefore $\alpha_n$ times its value during training.
Since no query outside subject $n$ may read $T_n$ under either regime (\cref{sec:method_apita}),
adding subjects does not directly expose subject $n$ to another subject's text,
and the same adapter can be applied without retraining for a particular subject count.

\section{Experiments}
\label{sec:experiments}

\subsection{Experimental Setup}
\label{sec:eval_setup}

\begin{table*}[!t]
  \centering
  \scriptsize
  \setlength{\tabcolsep}{5pt}
  \begin{tabular}{@{}lcccccc@{}}
    \toprule
    & \multicolumn{2}{c}{Protocol~C: joint editing}
    & \multicolumn{3}{c}{Protocol~D: two-dimensional control} \\
    \cmidrule(lr){2-3}\cmidrule(l){4-6}
    Method & Regional success $\uparrow$ & Worst-region $\Delta$CLIP $\uparrow$
      & Intended response $\uparrow$ & Semantic cross-talk $\downarrow$
      & LPIPS cross-talk $\downarrow$ \\
    \midrule
    (1) SliderEdit
      & 0.8231 & 0.0045 & 0.0307 & 0.6456 & 0.6597 \\
    (2) SliderEdit $+$ \apita{}
      & 0.8773 & 0.0122 & 0.0353 & \underline{0.2776} & \underline{0.1909} \\
    (3) \apita{} only
      & \textbf{0.9153} & \textbf{0.0282} & \multicolumn{3}{c}{not slider-capable} \\
    (4) Regional LoRA without \apita{}
      & 0.7577 & $-0.0015$ & \underline{0.0393} & 0.7146 & 0.9599 \\
    (5) \method{} (ours)
      & \underline{0.9099} & \underline{0.0251}
      & \textbf{0.0458} & \textbf{0.2211} & \textbf{0.1860} \\
    \bottomrule
  \end{tabular}
  \caption{\textbf{Protocols~C and D.}  Protocol~D applies only to
    slider-capable methods.  Best per column \textbf{bold}, second best
    \underline{underlined}.}
  \label{tab:protocol_cd}
\end{table*}

\paragraph{Evaluation benchmark.}
We evaluate on 128 multi-subject scenes containing 2--7 subjects per scene.
The benchmark provides 427 annotated subject--edit pairs spanning people,
animals, vehicles, furniture, and mixed compositions.  Each pair associates a
subject in the source image with a spatial annotation and a semantic edit
instruction.
The scenes are synthetic: we author a specification per scene and generate the
source image with \textsc{Flux.2}.  The benchmark is therefore disjoint from the
HQ-Edit training data by construction, and the generator differs from the
frozen editor, so no method is evaluated on images produced by the model it
edits with.  Spatial annotations are produced offline with
Grounding-DINO~\citep{liu2023grounding} and SAM2~\citep{ravi2024sam2}; these
annotators are not part of the editing pipeline, and at inference the
annotation is supplied by the user.  \Cref{sec:supp_bench} gives the full
construction.  Methods without an explicit spatial
interface instead receive positional language identifying the target, such as
``the leftmost dog'' or ``the second person from the right.''

\subsection{Quantitative Evaluation}
\label{sec:quant_results}

We call a method \emph{slider-capable} when it exposes a continuous,
per-subject edit control, and sweep every slider-capable method over
$\alpha\in\{1.0,0.8,\ldots,-0.8,-1.0\}$.

\paragraph{Compared methods.}
\begin{enumerate}[nosep,leftmargin=*]
  \item \textbf{SliderEdit.} The released Selective Token LoRA
  checkpoint~\citep{slideredit}, the multi-instruction variant, which scales
  instruction text spans but is given no mask, box, or routing.

  \item \textbf{SliderEdit $+$ \apita{}.} The same checkpoint with
  \apita{}~\citep{idattn} added at inference and each instruction block paired
  with its subject box, testing whether a naive combination works.

  \item \textbf{\apita{} only.} Routing and local instruction blocks, but no
  LoRA, so there is no learned strength control.

  \item \textbf{Regional LoRA without \apita{}.} The regional adapter trained
  with the masked suppression objective of \cref{sec:method_training}, but under ordinary full attention.

  \item \textbf{\method{} (ours).} The regional adapter and \apita{} routing,
  trained under the routing-matched objective of \cref{sec:method_training}.
\end{enumerate}
Two contrasts within this set carry the component analysis.  Methods 3, 4 and 5
isolate the two components, giving routing alone, the regional adapter alone,
and both together, which shows what either contributes by itself and whether
the two compound.  Methods 2 and 5 form the
second contrast, testing whether routing must be present while the regional
edit direction is learned or whether composing the two post hoc suffices.
% TODO: Decide whether Kontinuous Kontext is evaluated under Protocol A or is
% used only as methodological precedent for continuous-control evaluation.

\paragraph{Evaluation protocols.}
\begin{enumerate}[itemsep=3pt,topsep=3pt,parsep=0pt,leftmargin=*]
  \item \textbf{Protocol A: single-target editing.}
  We supply one subject-level instruction and sweep its control over the full
  range, with no instruction assigned to any other subject.

  \item \textbf{Protocol B: independent multi-control.}
  Every subject now carries an instruction, but we move one control and hold
  the rest at the neutral anchor $\alpha=1$.  This measures whether one control can move without dragging the
  others with it.

  \item \textbf{Protocol C: joint editing.}
  Every subject carries an instruction, and we generate one image with all
  regional controls at the nominal edit, $\alpha=0$, which we compare with the
  corresponding neutral output at $\alpha=1$.  This tests whether several
  requested edits remain effective when composed in the same image.

  \item \textbf{Protocol D: two-dimensional control.}
  We consider two-subject scenes and sweep both controls over the full grid
  $\alpha_A,\alpha_B\in\{-1,-\tfrac{2}{3},-\tfrac{1}{3},0,
  \tfrac{1}{3},\tfrac{2}{3},1\}$.  The resulting $7\times7$ grid
  forms a response surface: each row varies one subject while holding the
  other fixed, and each column does the converse.  This separates the response
  to a subject's own control from changes caused by the other control, exposing
  cross-talk that the joint endpoint in Protocol~C cannot reveal.
\end{enumerate}
Protocols~B and D require per-subject controls and are therefore run only on
the slider-capable methods.

\paragraph{Metrics.}
For Protocols~A and B we adopt the extrapolation and disentanglement metrics of
SliderEdit~\citep{slideredit}.  \emph{Extrapolation} is the maximum target
similarity attained over the control trajectory, measured with
CLIP~\citep{radford2021clip} and SigLIP~\citep{zhai2023siglip}, and
\emph{disentanglement} is the LPIPS~\citep{zhang2018lpips},
DINOv2~\citep{oquab2023dinov2} and ArcFace~\citep{deng2019arcface} distance
between each edited output and the source image.  We do not adopt their continuity score, which by construction cannot
distinguish a wide, well-ordered sweep from a narrow or noisy one;
\cref{sec:supp_continuity} gives the argument.  In its place we report two
diagnostics of the trajectory itself.  \emph{Direction correlation} is the
Spearman rank correlation between $\alpha$ and the target similarity.
\emph{Mean range} is the average gap between the highest and lowest similarity
a trajectory reaches, which reports how far the control actually moves the
target.  Target similarity for
these diagnostics is measured between the edit text and a crop of the edited
image formed from the subject bounding box, independently with CLIP and SigLIP.

Protocol~C scores each region by the CLIP similarity between a crop of the
subject and its edit description, and measures how much that similarity
improves from the neutral output to the nominal edit.  \emph{Regional success
rate} is the fraction of regions that show an edit, and \emph{worst-region
$\Delta$CLIP} is the smallest improvement among the subjects.  Protocol~D moves
one control by a single step and records the size of each region's change per
unit of control.  Moving a subject's own control should change that subject,
and the average of those changes is the \emph{intended response}.  Moving the
other subject's control should change nothing, so \emph{semantic cross-talk}
reports that unintended change as a fraction of the intended one, averaged over
the two subjects, and \emph{perceptual cross-talk} repeats this with regional
LPIPS in place of the CLIP score.
\Cref{sec:supp_metrics} gives the formal definitions of every metric above.

\paragraph{Implementation details.}
All five methods use FLUX.1-Kontext-dev~\citep{flux_kontext} as the frozen base
editor and generate at $1024\times1024$ resolution with 28 denoising steps,
guidance scale 3.5, and seed 0.  Only prompt syntax and spatial inputs vary,
following each method's interface.  The two trained methods add a rank-$16$
LoRA to the frozen transformer and train it for $10{,}000$ steps on
$8\times$H200 GPUs over $22{,}482$ examples derived from
HQ-Edit~\citep{hqedit}, each pairing a source image with one subject mask and
one instruction; \cref{sec:supp_impl} gives the full configuration.

\paragraph{Edit strength and preservation.}
Across Protocols~A and B, \method{} attains both the highest extrapolation and
the lowest disentanglement on every metric and both encoders
(\cref{tab:slideredit_metrics_available}).  These two normally trade off, since
a method can preserve the source simply by editing less; here the strongest
preservation coincides with the strongest target alignment, so it does not come
from weaker editing.

\paragraph{Continuous control.}
\method{} also gives the most reliably ordered response and the widest semantic
range, under both protocols and both encoders.

\paragraph{Joint editing.}
Under Protocol~C, routing is what allows several requested edits to land at
once (\cref{tab:protocol_cd}).  \apita{} alone gives the best coverage but
exposes no continuous control, and \method{} comes close to matching it while
retaining an independent slider per subject, so continuous control costs little
coverage.

\paragraph{Two-dimensional control.}
Protocol~D separates the components most clearly and shows that neither
substitutes for the other.  The regional adapter without routing produces a
strong intended response but the worst cross-talk in the table, worse even than
unmodified SliderEdit: the control moves its target, but its effect is not
confined there, which is what a spatially entangled edit direction predicts.
Post-hoc composition fails in the opposite way, containing the edit well while
responding weakly.  Only training the adapter under the routing it will be used
with achieves both, giving the strongest intended response together with the
lowest cross-talk on both measures, though its perceptual margin over post-hoc
composition remains small.  Routing supplies the spatial isolation;
routing-aware training supplies the control.

\subsection{Qualitative Results}
\label{sec:qual_results}

\Cref{fig:qual} shows four multi-subject scenes.  Each block places the source
image on the left, with the {\color{sub1}red} and {\color{sub2}blue} boxes
marking the spatial annotations supplied to the model and the corresponding
instructions printed beneath it.  The three rows isolate the controls: the
first sweeps subject~1 while subject~2 is held at the neutral anchor
$\alpha=1$, the second sweeps subject~2 while subject~1 is held neutral, and
the third moves both.  Within a row the edit strengthens from left to right.
The final block is the exception, applying one instruction to both annotated
regions rather than a separate instruction to each. We encourage referring to \cref{fig:supp_fourway} and \cref{fig:supp_fourway_vases} in the appendix for examples where more than two objects are edited.

% !TeX root = ../main.tex
\section{Conclusion}
\label{sec:conclusion}

We presented \method{}, a framework for region-aware continuous image editing that edits multiple subjects
in one image at independently chosen strengths within a single forward pass.
By replacing the adapter's scalar coefficient with a per-token vector aligned to the token partition,
and training that adapter under the same instance-disentangled routing used at inference,
we obtain region-local edit directions whose rescaling gives each subject its own continuous control.

\clearpage
\bibliographystyle{assets/plainnat}
\bibliography{references}

\clearpage
\beginappendix
\crefalias{section}{appendix}
\crefname{appendix}{Appendix}{Appendices}
\Crefname{appendix}{Appendix}{Appendices}
% !TeX root = ../main.tex
\section{Extended Related Work}
\label{sec:supp_related}

This section expands the comparison summarized in \cref{sec:related}.

\paragraph{Instruction-following image editing.}
InstructPix2Pix~\citep{brooks2023instructpix2pix} established the paradigm by training on synthetically generated edit triplets,
MagicBrush~\citep{zhang2024magicbrush} improved data quality with manually annotated pairs,
and Emu Edit~\citep{sheynin2024emu} decomposed editing into recognition and generation sub-tasks.
\flux{}~\citep{flux_kontext} concatenates source and noise latents inside an MMDiT backbone.

\paragraph{Spatial control.}
Prompt-to-Prompt~\citep{hertz2022prompt} edits by swapping cross-attention maps between prompts,
and MasaCtrl~\citep{cao2023masactrl} obtains consistent synthesis by sharing mutual self-attention keys and values.
GLIGEN~\citep{li2023gligen} injects bounding box or keypoint conditions through gated self-attention,
but targets generation rather than editing.

\paragraph{Continuous edit-strength control.}
NumeriKontrol~\citep{numerikontrol} conditions on explicit numeric descriptors written into the prompt,
which gives numeric control over an attribute but no spatial addressing.

\paragraph{Multi-subject and localized editing.}
Differential Diffusion~\citep{diffdiff} assigns each pixel or region its own strength through a change map.
Because the map scales the per-pixel noising schedule under a single global prompt,
it modulates how much a region changes rather than applying distinct edits to different subjects,
and it provides no attention-level isolation.
Layered Diffusion Brushes~\citep{layeredbrushes} exposes a per-layer strength that scales the Gaussian noise injected into a user-drawn region.
That control is a regeneration amount coupled to the injection timestep,
rather than a disentangled slider interpolating a single edit from source to full application.
Both rely on manually specified regions and allow regions to interact without isolation.

\paragraph{Cross-subject interference.}
Interference between subjects is well documented.
Bounded Attention~\citep{dahary2024bounded} attributes it to semantic leakage between subjects in the attention layers,
and text-embedding leakage has been shown to spill edits onto unintended regions~\citep{mun2025leakage}.
These studies motivate a dedicated isolation mechanism rather than reliance on language alone.

\section{Why We Do Not Report Semantic Continuity}
\label{sec:supp_continuity}

SliderEdit~\citep{slideredit} summarizes the continuousness of a control trajectory
with a continuity score.  Given the similarity sequence $s_1,\dots,s_\delta$
recorded over the $\alpha$ sweep, the interval $[\min_i s_i,\max_i s_i]$ is
divided into $\delta$ bins, the number of scores falling in each bin is counted,
and the reported value is $(\chi^2/\mathrm{dof})^{-1}$ against the uniform
expectation of one score per bin.  Higher values indicate a more evenly
traversed range.

Two normalizations in that construction discard the information a slider is
meant to provide.  Binning counts is invariant to order, so a trajectory that
oscillates scores exactly as well as one that rises monotonically.  Binning
between the observed minimum and maximum is invariant to scale, so a trajectory
whose similarity moves by $0.005$ is stretched to fill the same bins as one that
moves by $0.30$.  A control that barely responds therefore scores highly:
measurement noise is close to uniformly distributed by construction, which is
exactly what the statistic rewards.

We therefore report the trajectory diagnostics of
\cref{tab:slideredit_metrics_available} instead: Spearman correlation restores
sensitivity to order, and mean range restores sensitivity to scale.

\section{Aggregation}
\label{sec:supp_aggregation}

Multiple subject--edit pairs derive from the same source image and are therefore
correlated.  Unless stated otherwise, we first average subject-level measurements
within each source image and then average equally over source images, rather than
treating the 427 subject-level pairs as independent observations.  The trajectory
diagnostics of \cref{tab:slideredit_metrics_available} instead take one
source--region--instruction trajectory as their unit and average over all 427
trajectories.

\section{Formal Metric Definitions}
\label{sec:supp_metrics}

Let $u=1-\alpha$ denote edit strength, so that $u=0$ is the exact neutral
output and $u=1$ is the nominal edit.  Let $I(u)$ be the image generated with
all regional controls set to $u$, let $p_n$ be the target-state description
corresponding to instruction $e_n$, and let $\mathrm{crop}(I,M_n)$ be the
masked crop of region $n$.  The regional score is
\begin{equation}
  q_n(u)=\operatorname{sim}_{\mathrm{CLIP}}
  \bigl(\mathrm{crop}(I(u),M_n),\,p_n\bigr).
  \label{eq:supp_regional_score}
\end{equation}

\paragraph{Protocol C.}
At the nominal edit we compute
\begin{equation}
\begin{aligned}
  \Delta_n &= q_n(1)-q_n(0), \\
  \mathrm{Succ} &= \frac{1}{N}\sum_{n=1}^{N}\mathbf{1}[\Delta_n>0],
  \qquad \mathrm{Worst}=\min_n \Delta_n .
\end{aligned}
\label{eq:supp_protocol_c}
\end{equation}
Regional success rate counts how often a requested edit improves its own
target score, while worst-region gain exposes a scene in which composition
sacrifices even one subject.

\paragraph{Protocol D.}
Each cell of the $7\times7$ grid is an image $I(u_1,u_2)$, and the two regional
score surfaces $q_1(u_1,u_2)$ and $q_2(u_1,u_2)$ follow
\cref{eq:supp_regional_score}.  Comparing cells that differ by one step in a
single control gives $6\times7=42$ adjacent pairs per control direction.  For
region $i$ and control $j$,
\begin{equation}
  S_{i\leftarrow j}=\operatorname{mean}
  \frac{\bigl|\Delta_j q_i\bigr|}{\Delta u},
  \label{eq:supp_response}
\end{equation}
where $\Delta_j q_i$ is the change in $q_i$ across one step of control $j$
and the mean runs over those 42 pairs.  We report
\begin{equation}
\begin{aligned}
  \mathrm{Resp}_{\mathrm{int}} &= \tfrac{1}{2}
    \bigl(S_{1\leftarrow1}+S_{2\leftarrow2}\bigr), \\
  \mathrm{Cross}_{\mathrm{sem}} &= \tfrac{1}{2}\Bigl(
    \tfrac{S_{1\leftarrow2}}{S_{1\leftarrow1}}+
    \tfrac{S_{2\leftarrow1}}{S_{2\leftarrow2}}\Bigr).
\end{aligned}
\label{eq:supp_crosstalk}
\end{equation}
Perceptual cross-talk replaces the change in CLIP score with the LPIPS
distance between regional crops of adjacent generated images, so
$P_{i\leftarrow j}$ substitutes for $S_{i\leftarrow j}$ in
\cref{eq:supp_crosstalk}.  Unlike the source-referenced disentanglement
measure of SliderEdit~\citep{slideredit}, both cross-talk terms compare
neighbouring generated outputs rather than each output against the source.
Because \cref{eq:supp_response} takes absolute differences and therefore
carries no sign, we separately verify that the mean signed response is
positive.

\section{Benchmark Construction}
\label{sec:supp_bench}

The 128 evaluation scenes are curated rather than sampled from an existing
corpus.  For each scene we authored a specification giving a source-image
generation prompt, the intended number and type of subjects, and one edit
instruction per subject.  Source candidates were generated at
$1024\times1024$ with \textsc{Flux.2}, using 28 denoising steps, guidance
scale $3.5$, and typically four fixed seeds per scene.  We inspected the
candidates and kept images showing the intended subject count; scenes that
failed this check were regenerated from revised prompts.  Note that the generator differs from the
frozen editor used at inference, \flux{}, so no method is evaluated on images
produced by the model it edits with.

Grounding-DINO~\citep{liu2023grounding} then detected the requested subjects in
each retained image.  Detections were de-duplicated and ordered left to right,
which fixes the subject indices used throughout the protocols, and each box was
passed to SAM2~\citep{ravi2024sam2} for a subject silhouette.  Every scene
stores the source image, the per-subject boxes and masks, the detection
confidence and mask area, and the ordered per-subject instructions.

Instructions follow the prompt construction of SliderEdit~\citep{slideredit},
extended by LLM-assisted generation and manual curation so that each subject
receives a single atomic edit that can be verified visually.

\section{Implementation Details}
\label{sec:supp_impl}

\paragraph{Training data.}
Training examples are derived from HQ-Edit~\citep{hqedit}.  We decode all
$98{,}675$ source images together with their original edit instructions.  Each
instruction is passed to a vision--language model~\citep{qwen3vl}, which assigns
it one of four types, \textsc{modify}, \textsc{add}, \textsc{remove} or
\textsc{global}, and names the primary target noun; the instruction text itself
is left unchanged.  That noun prompts Grounding-DINO~\citep{liu2023grounding},
and the resulting box is passed to SAM2~\citep{ravi2024sam2} to obtain a mask.
We discard examples for which no usable mask is produced, and those whose
instruction requests a change of pose, movement, size, rotation or camera
viewpoint, as these do not correspond to a confined image region.  The
resulting manifest holds $23{,}591$ examples; removing instances whose mask
covers less than $3\%$ of the image leaves $22{,}482$ training examples
spanning $21{,}962$ distinct instructions.  Rows are shuffled during training,
but instructions are used as written and are never resampled, recombined or
synthesised.

The VAE, the T5 and CLIP text encoders, and the scheduler are frozen wherever
applicable, and we train only a regional adapter: a rank-$16$ LoRA without
dropout wrapping the linear layers of the transformer, stored in fp32 while the
frozen backbone computes in bf16.

The per-region write gate projects each subject's mask onto the transformer's
token grid.  The VAE downsamples by $8\times$ and \flux{} packs $2\times2$
latent neighbourhoods into one token, so each token cell spans $16\times16$
pixels and a $1024\times1024$ input yields a $64\times64$ grid.  The gate takes
the max over each cell rather than a coverage threshold, which keeps thin
structures alive at the cost of dilating each gate by at most one token.  The
resulting per-token scale is applied stream-aware: text entries are sliced into
the dual-stream text modules, image entries into the dual-stream image modules,
and the full vector into single-stream blocks.

The text sequence uses a $512$-token T5 encoding truncated to the first $200$
positions per instruction block.  Combined with $4096$ noise and $4096$
source-image tokens, the joint sequence is $8{,}392$ tokens for a single-region
training example.

Both adapters are trained for $10{,}000$ steps on $8\times$H200 GPUs under data
parallelism with one example per rank (effective batch size $8$), using AdamW
($\text{lr}=7\times10^{-5}$, weight decay $10^{-5}$), a cosine schedule with
$500$ warmup steps, gradient clipping at $1.0$, gradient checkpointing,
logit-normal timestep sampling ($\mu=0,\sigma=1$), guidance embedding $3.5$, and
no SNR reweighting.

\section{Four-Subject Control}
\label{sec:supp_fourway}

\begin{figure*}[!t]
  \centering
  \qualquad[Chair]{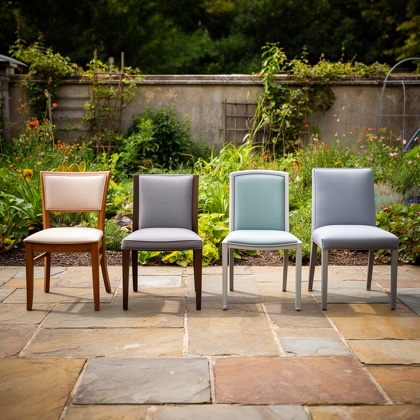}{%
    {\color{sub1}\emph{turn the chair into an ornate gold throne}}\par\vspace{2pt}%
    {\color{sub2}\emph{make the chair look old and weathered with torn fabric}}\par\vspace{2pt}%
    {\color{sub3}\emph{turn the chair into an acrylic chair}}\par\vspace{2pt}%
    {\color{sub4}\emph{make the chair a wooden chair}}}
  \caption{\textbf{Four independent controls in one scene.}  The input carries
    four spatial annotations, drawn in {\color{sub1}red}, {\color{sub2}blue},
    {\color{sub3}green} and {\color{sub4}orange}, each with its own
    instruction.  Each of the first four rows moves one chair's control while
    the other three are held at the neutral anchor $\alpha=1$; the last row
    moves all four together.  Every column shows an edited output; the
    unedited scene is the input on the left.}
  \label{fig:supp_fourway}
\end{figure*}

\begin{figure*}[!t]
  \centering
  \qualquad[Vase]{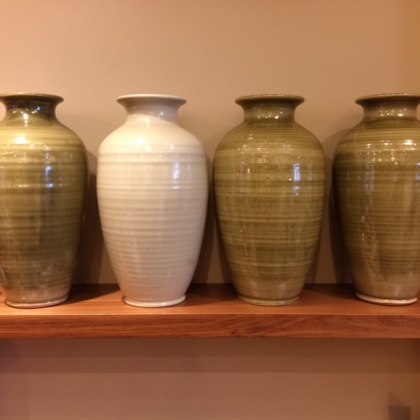}{%
    {\color{sub1}\emph{turn the vase into an ornate golden vase with jewels}}\par\vspace{2pt}%
    {\color{sub2}\emph{make the vase look ancient with cracks and patina}}\par\vspace{2pt}%
    {\color{sub3}\emph{turn the vase into a crystal glass vase}}\par\vspace{2pt}%
    {\color{sub4}\emph{make the vase appear to be carved from rough weathered stone}}}
  \caption{\textbf{Four independent controls on a second scene.}  Each of the
    first four rows moves one vase's control while the other three are held at
    the neutral anchor $\alpha=1$, and the last row moves all four together.
    The first column is the unedited output at $\alpha=1$.}
  \label{fig:supp_fourway_vases}
\end{figure*}

The vases of \cref{fig:supp_fourway_vases} are the case positional language
handles worst: the four targets are visually similar and differ mainly by
position, which is where an explicit spatial annotation matters most.

\end{document}